# A SURVEY ON EYE-GAZE TRACKING TECHNIQUES

H.R. Chennamma

Department of MCA, Sri Jayachamarajendra College of Engineering,
Mysore, Karnataka, INDIA
anuamruthesh@gmail.com

Xiaohui Yuan

University of North Texas,
Denton, Texas, USA
xiaohui.yuan@unt.edu

**Abstract**

Study of eye-movement is being employed in Human Computer Interaction (HCI) research. Eye - gaze tracking is one of the most challenging problems in the area of computer vision. The goal of this paper is to present a review of latest research in this continued growth of remote eye-gaze tracking. This overview includes the basic definitions and terminologies, recent advances in the field and finally the need of future development in the field.

## 1. Introduction

Face is the index of mind and eyes are the window to the soul. Eye movements provide a rich and informative window into a person's thought and intentions. Thus the study of eye movement may determine what people are thinking based on where they are looking. Eye tracking is the measurement of eye movement/activity and gaze (point of regard) tracking is the analysis of eye tracking data with respect to the head/visual scene. Researches of this field often use the terms eye-tracking, gaze-tracking or eye-gaze tracking interchangeably. Eye tracking is mostly used in the applications like drowsiness detection [Picot *et al.* (2010)], diagnosis of various clinical conditions or even iris recognition [Xu *et al.* (2006)]. But gaze tracking methods can be used in all the ways that we use our eyes, to name a few, eye typing for physically disabled [Majaranta *et al.* (2002)], Cognitive and behavioural therapy [Riquier *et al.*(2006)], visual search [Greene *et al.* (2001)], marketing/advertising [Rayner *et al.* (2001)], neuroscience [Snodderly *et al.* (2001)], Psychology [Rayner (1998)] and Human Computer Interaction (HCI) [Goldberg *et al.* (2002); Jacob (1990) ].

Usually the integration of eye and head position is used to compute the location of the gaze in the visual scene. Simple eye trackers report only the direction of the gaze relative to the head (with head-mounted system, electrodes, sceleral coils) or for a fixed position of the eyeball (systems which require a head fixation). Such eye tracking systems are referred as intrusive or invasive systems because some special contacting devices are attached to the skin or eye to catch the user's gaze [Duchowsky (2007)]. The systems which do not have any physical contact with the user and the eye tracker apparatus are referred as non-intrusive systems or remote systems [Morimoto and Mimica (2005)].

## 2. Methods of Eye Tracking

A method of recording eye position and movements is called oculography. There are four different methods to track the motion of the eyes [COGAIN (2005)].

### 2.1. Electro-Oculography

In this method, sensors are attached at the skin around the eyes to measure an electric field exists when eyes rotate. By recording small differences in the skin potential around the eye, the position of the eye can be estimated. By carefully placing electrodes, it is possible to separately record horizontal and vertical movements. However, the signal can change when there is no eye movement. This technique is not well-suited for everyday use, since it requires the close contact of electrodes to the user but is still frequently used by clinicians. However, it is a cheap, easy and invasive method of recording large eye movements. The big advantage of this method is its ability to detect eye movements even when the eye is closed, e.g. while sleeping [Mazo *et al.* (2002)]. The projects called MONEOG [http://www.metrovision.fr], from Metro Vision Systems and Eagle Eyes [http://www.bc.edu/eagleeyes] from Opportunity Foundation of America have used the method of electro-oculography successfully for eye-gaze tracking. The Eagle Eyes have been helping people with severe physical disabilities to control the computer by moving only their eyes.

## 2.2. Sceleral Search Coils

When a coil of wire moves in a magnetic field, the field induces a voltage in the coil. If the coil is attached to the eye, then a signal of eye position will be produced. In order to measure human eye movements, small coils of wire are embedded in a modified contact lens. This is inserted into the eye after local anesthetics has been introduced. An integrated mirror in the contact lens allows measuring reflected light. Alternatively, an integrated coil in the contact lens allows detecting the coil's orientation in a magnetic field. The advantage of such a method is the high accuracy and the nearly unlimited resolution in time. Its disadvantage is that it is an invasive method, requiring something to be placed into the eyes. To the best of our knowledge, this method of eye tracking has not been used for HCI by gaze, so far. This method is mostly used in medical and psychological research. Chronos Vision [http://www.chronos-vision.de] and Skalar Medical [http://www.nzbri.org/research/labs/eyelab/] have used scleral search coils method for eye tracking relative to the head position.

## 2.3. Infrared Oculography

The infrared oculography measures intensity of reflected infrared light. In this eye tracking method, eye is illuminated by infrared light which is reflected by the scelera. The difference between the amounts of IR light reflected back from the eye surface carries the information about the eye position changes. The light source and sensors can be placed on spherical glasses. Hence it is an invasive method. The infrared oculography has less noise than electro-oculography, but is more sensitive on changes of external light tension. The main disadvantage of this method is that it can measure eye movement only for about ±35 degrees along the horizontal axis and ±20 degrees along the vertical axis. Yet an application can be found in JAZZ-novo and saccadometer research systems [http://www.ober-consulting.com]. These systems are designed to measure eye movements during Magnetic Resonance Imaging (MRI) examination. The advantages include ability to measure eye movements in darkness. Infrared oculography is being used in gaze interaction by making use of image processing software. There are three categories of infrared oculography which use: the corneal reflection, the Purkinje images and the track of the pupil. These principles have been exploited in a number of commercially available eye trackers viz. Intelligaze IG-30 [http://www.alea-technologies.com], EyeMax System [http://www.dynavoxtech.com], EyeTech Digital Systems [http://www.eyetechds.com] and SeeTech[http://www.see-tech.de].

## 2.4. Video Oculography

Video-based eye tracking is the most widely used method in commercial eye trackers. Until recently, the eye-gaze tracking was a very complex and expensive task limited for only laboratory research. However, rapid technological advancements (increased processor speed, advanced digital video processing) have lowered the cost and dramatically increased the efficiency of eye - gaze tracking equipment. Video oculography make use of single or multiple cameras to determine the movement of eye using the information obtained from the images captured.

Video-based eye tracking systems may be invasive or non-invasive. Each category again splits into two other categories depending on the kind of light used: visible light or infrared light. Invasive systems or head mounted systems are commonly composed of one or more cameras [Duchowsky (2007)]. Non-invasive or remote systems are the most exciting subject of Human Computer Interactions (HCI) [Huchuan *et al.* (2012); Morimoto and Mimica (2005)]. In this paper, we are concentrating on video-based remote eye tracking systems. It is surprising to find the wide variety of gaze tracking systems which are used with the same purpose, that is, to detect the point of gaze [Hansen and Ji. (2010); Böhme *et al.* (2006); Orman *et al.* (2011); Černy, M. (2011); Mohamed *et al.* (2008)]. However, their basis seems to be the same; the image of the eye captured by the camera will change when eye rotates or translates in 3D space. The remote eye tracking systems that appeared in the literature can be grouped into; single-camera eye tracker and multi-camera eye tracker. The following section focus on hardware setup of the eye tracker system rather than some cumbersome mathematical details.

### 2.4.1. Single Camera Eye Tracker

Most video-based eye trackers work by illuminating the eye with an infrared light source. This light produces a glint on the cornea of the eye and is called as corneal reflection. In most of the existing work, glint has been used as the reference point for gaze estimation. The pupil-glint difference vector remains constant when the eye or the head moves. The glint will clearly change location when the head moves, but it is less obvious that the glint shifts position when changing gaze direction. We find in the very beginning work by Merchant et al. in 1974 employing a single camera, a collection of mirrors and a single illumination source to produce the desired effect [Merchant *et al.* (1974)]. Several commercial systems base their technology on one camera and one infrared light such as the trackers from LC [http://www.eyegaze.com] or ASL [http://www.a-s-l.com].

Some systems incorporate a second lighting, as the one from Eyetech, [*http://www.eyetechds.com*]. Yasuk Sugano et al. proposed a gaze estimation system with a single camera mounted on the monitor by using incremental learning method. This system also estimates the head pose of a person by using a 3D rigid facial mesh [Sugano *et al.* (2008)]. Ohno et al developed a single camera system with a single glint [Ohno *et al.*, (2002)]. Matsumoto et al. proposed a system which uses a single stereo system to compute the 3D head pose and estimate the 3D position of the eyeball [Matsumoto *et al.* (2000)]. A similar approach is also proposed by Wang and Sung (2002). Nitschke et al. proposed an eye pose estimation model by using single camera and display monitor is used as a light source [Christian *et al.* (2011)]. Paul Smith et al. described a system for monitoring head/eye motion for driver alertness with a single camera [Paul Smith *et al.* (2000)]. Hirotake et al. proposed a remote gaze estimation method based on facial-feature tracking using a single video camera [Yamazoe *et al.* (2008)]. Wang et al. introduced eye-gaze estimation method just by using one camera based on iris detection [Wang *et al.* (2003)]. Chi Jian-nan et al proposed a pupil tracking method based on particle filtering with an active infrared source and a camera [Chi *et al.* (2011)]. Laura Sesma et al. proposed a gaze tracking system based on a web cam and without infrared light is a searched goal to broaden the applications of eye tracking systems [Sesma *et al.* (2012)]. Xiao-Hui Yang et al. employed only one camera and four IR light sources and used the gray distribution of the video frame to obtain the corneal glints and pupil centre [Yang *et al.* (2012)].

The main difficulty with the above fixed single camera systems is the limited field of view required to capture sufficiently high resolution images. By adding multiple light sources to the setup will provide better results than with the single source. The first single camera remote eye tracker with high accuracy (about 1 degree) and good tolerance to user movement was a commercial system (Tobii: *http://www.tobii.se*] but implementation details have not been available. Several academic groups have built single camera systems [Hennessey *et al.* (2006); Guestrin and Eizenman (2006); Meyer *et al.* (2006)]. Ohno et al. improved their system with two light sources and a single camera [Ohno (2006)]. Morimoto et al. introduced a method for computing the 3D position of an eye and its gaze direction from a single camera and at least two light sources [Morimoto *et al.* (2002)]. The authors argue that it allows for free head motion. Tomono et al. developed a real time imaging systems composed of a single camera with 3 CCDs and two light sources [Tomono *et al.* (1989)]. A camera is located slightly below the centre of the screen. Four light sources are placed at the corners of a planar surface (screen) to be able to compare and cross-ratio methods [Hansen *et al.* (2010); Flavio *et al.* (2012)]. Xiaohui et al. proposed an intelligent control scheme for remote gaze tracking which includes an ordinary resolution camera and four near infrared light sources [Xiaohui *et al.* (2010)].

*2.4.2. Multi-camera Eye Tracker*

A large field of view is required to allow for free head motion, but a limited field of view is needed to capture sufficiently high-resolution eye images to provide reliable gaze estimates. Multiple cameras are utilized to achieve these goals either through wide-angle lens cameras or movable narrow-angle lens cameras. Multiple camera systems in the literature use either separate cameras for each eye or use one camera for head location tracking to compensate for head pose changes. Then combine the information of all the cameras to estimate gaze point. Zhu et al. proposed an eye gaze tracking system in which two video cameras are mounted under the monitor screen and an IR illuminator is mounted in the front of one camera to produce the glint in the eye image. Therefore, the pupil-glint vector can be extracted from the captured eye images. In addition, both cameras are calibrated to form a stereo vision system so that the 3D coordinate of the pupil center can be computed. The computed 3D pupil center will concatenate with the extracted 2D pupil-glint vector to serve as the input for the gaze mapping function [Zhu *et al.* (2006)]. Baymer and Flickner present a system with four cameras: two stereo wide angle cameras and two stereo narrow field of view cameras. Two narrow field cameras that are positioned near the lower monitor corners capture high resolution images of the eye for gaze tracking. Due to the narrow field of view, quick head motions would outpace pan-tilt heads. Thus, pan and tilt are controlled using rotating mirrors on high performance galvos. Two wide angle systems are located just below the center bottom of the monitor screen. The stereo baseline is oriented vertically since this optimizes stereo matching on the central facial features, which are predominantly horizontal edges [Beymer and Flickner (2003)]. Similar system was proposed by Brolly et al, but unlike their system, they use a single narrow field eye camera instead of two [Brolly and Mulligan (2004)]. Ohno and Mukawa implemented a free-head gaze tracking system. The eye positioning unit has a stereo camera set which consists of two NTSC cameras. They are placed on a display monitor. The gaze tracking unit has a near-infrared sensitive NTSC camera placed on a pan-tilt stand. A near-infrared LED array is also placed under the camera. However, the measurable area can be expanded by changing the stereo cameras' focal lengths and the convergence angle [Ohno. and Mukawa (2004)]. Shih et al. proposed a system that consists of a pair of stereo cameras and three IR LEDs [Shih and Liu (2004)]. Ke Zhang et al. developed a simplified 3D gaze tracking technology with a pair of stereo cameras and two point light sources [Ke *et al.* (2010)]. Yoo and Chung employed five infrared lights and two cameras [Yoo and Chung, (2005)]. Reale et al. present a two-camera system that detects the face from a fixed, wide-angle camera, estimates a rough location for the eye region using an eye detector based on topographic features, and directs another active pan-tilt-zoom camera to focus on this eye region [Hung and Yin (2010)].

### 3. Methods of Gaze Tracking

Video oculography systems obtain information from one or more cameras (image data). The first step is to detect the eye location in the image. Based on the information obtained from the eye region and possibly head pose, the direction of gaze can be estimated. The most important parts of human eye are: the pupil – the aperture that lets light into the eye, the iris – the colored muscle group that controls the diameter of the pupil and the scelera – the white protective tissue that covers the remainder of the eye. Eye detection and tracking remains a very challenging task due to several unique issues, including illumination, viewing angle, occlusion of the eye, head pose etc. Two types of imaging processes are commonly used in video-based eye tracking: visible and infrared spectrum imaging. Infrared eye tracking typically utilizes either bright pupil or dark-pupil technique [Morimoto *et al.* (2002)]. In this paper, we focus on gaze estimation methods based on analysis of the image data. These methods are broadly grouped into feature-based and appearance-based gaze estimation [Hansen and Ji. (2010)].

**3.1. Feature-based Gaze Estimation**

Feature-based methods explore the characteristics of the human eye to identify a set of distinctive features of the eyes like contours (limbus and pupil contour), eye comers and cornea reflections are the common features used for gaze estimation. The aim of feature-based methods is to identify informative local features of the eye that are generally less sensitive to variations in illumination and viewpoint [Iannizzotto and La Rosa (2011)]. These systems have performance issues in outdoors or under strong ambient light. In addition, the accuracy of gaze estimation decreases when accurate iris and pupil features are not available. There are two types of feature-based approaches exists [Hansen and Ji (2010)]: model-based (geometric) and interpolation-based (regression-based).

*3.1.1. Model-based approaches*

Model-based approaches use an explicit geometric model of the eye to estimate 3D gaze direction vector. Most 3D model-based (or geometric) approaches rely on metric information and thus require camera calibration and a global geometric model (external to the eye) of light sources, camera and monitor position and orientation. Most of the model-based method follows a common strategy: first the optical axis of the eye is reconstructed in 3D: the visual axis is reconstructed next: finally the point of gaze is estimated by intersecting the visual axis with the scene geometry. Reconstruction of the optical axis is done by estimation of the cornea and pupil centre. By defining the gaze direction vector and integrating it with information about the objects in the scene, the point of gaze is estimated [Hansen and Ji. (2010)]. For 3D model based approaches, gaze directions are estimated as a vector from the eyeball centre to the iris centre [Yamazoe et al. (2008); Taba (2012); Sigut and Sidha (2011); Yang et al. (2012); Hung and Yin (2010); Nagamatsu, et al. (2010); Model and Eizenman (2010)].

*3.1.2. Interpolation-based approaches*

These methods assume the mapping from image features to gaze co-ordinates (2D or 3D) have a particular parametric form such as a polynomial or a nonparametric form as in neural networks. Since the use of a simple linear mapping function in the first video-based eye tracker [Merchant *et al.* (1974)], polynomial expressions have become one of the most popular mapping techniques [Brolly and Mulligan (2004); Cerrolaza *et al.* (2012); Morimoto and Mimica (2005); Cerrolaza, *et al.* (2012)]. Interpolation-based methods avoid explicitly modeling the geometry and physiology of the human eye but instead describe the gazed point as a generic function of image features. Calibration data are used to calculate the unknown coefficients of the mapping function by means of a numerical fitting process, such as multiple linear regressions. As an alternative to parametric expressions, neural network-based eye trackers [Baluja and Pomerleau (1994); Demjen *et al.* (2011); Torricelli *et al.* (2008)] assume a nonparametric form to implement the mapping from image features to gaze coordinates. In these approaches, the gaze tracking is done by extracting the coordinates of certain facial points and sending them through a trained neural network, whose output is coordinates of the point where the user is looking at.

**3.2. Appearance-based Gaze Estimation**

Appearance-based methods detect and track eyes directly based on the photometric appearance. Appearance-based techniques use image content to estimate gaze direction by mapping image data to screen coordinates [Javier *et al.* (2009); Lu *et al.* (2011)]. The major appearance-based methods [Sheela and Vijaya (2011)] are based on morphable model [Rikert and Jones (1998)], gray scale unit images [Yang (2012)], appearance manifold [Kar-Han *et al.* (2002)], Gaussian interpolation [Sugano *et al.* (2012)] and cross-ratio [Flavio *et al.*

(2012)]. Appearance-based methods typically do not require calibration of cameras and geometry data since the mapping is made directly on the image contents.

## 4. Summary and Conclusion

With the introduction of different methods of eye tracking, we have presented a review of non-contacting video-based gaze tracking. The main intention of this paper is to give a review of latest growth in non-contacting video-based gaze tracking. Even though, the eye-gaze tracking has a history of 100 years of research, it has not been standardized. Future developments in eye tracking need to centre on standardizing what eye movement metrics are used, how they are referred to, and how they should be interpreted in the context of interface design [Poole *et al.* (2004)]. For example, no standard yet exists for the minimum duration of a fixation. The intrusiveness of equipment should be decreased to make users feel more comfortable. The robustness and accuracy of data capture needs to be increased and eye-gaze tracking systems need to become cheaper in order to make them a viable usability tool for smaller commercial agencies and research labs.